%% file: main.tex
\documentclass{article}
\PassOptionsToPackage{authoryear,round}{natbib}
\usepackage[preprint]{neurips_2026}
\makeatletter
\renewcommand{\@notice}{}
\makeatother
\usepackage[utf8]{inputenc}
\usepackage[T1]{fontenc}
\usepackage{hyperref,url,booktabs,amsfonts,nicefrac,microtype,xcolor,graphicx,multirow,float,placeins}
\usepackage{amsmath}
\hypersetup{
  colorlinks=false,
  pdfborder={0 0 0},
  pdftitle={MedAdaptAgent: An Automatic Harness for Adapting Medical Imaging Models to New Sites},
  pdfauthor={Hanxue Gu, Zhonghao Liu, Xingrui Gu, Kang Wang, Yang Yang},
  hypertexnames=false
}

\newcommand{\systemname}{\textit{MedAdaptAgent}}
\newcommand{\onboardingname}{\textit{Onboarding Analyzer}}
\newcommand{\plannername}{\textit{Adaptation Planner}}
\newcommand{\harnessname}{\textit{MedAdaptAgent} harness}
\newcommand{\gatename}{\textit{Evidence Gate}}

\title{\systemname{}: An Automatic Harness for Adapting Medical Imaging Models to New Sites}
\author{%
Hanxue Gu$^{1}$ \quad Zhonghao Liu$^{1}$ \quad Xingrui Gu$^{2}$ \\
\textbf{Kang Wang$^{1}$ \quad Yang Yang$^{1}$} \\
$^{1}$University of California, San Francisco \\
$^{2}$University of California, Berkeley
}

\begin{document}
\raggedbottom
\maketitle

\begin{abstract}
Reusing pretrained medical-imaging models offers an alternative to training from scratch. A model trained for the same task may perform poorly at a new hospital because imaging protocols and preprocessing differ. Effective adaptation depends on the model architecture, domain shift, and available labels and compute. Adaptation can improve scores on a small labeled set yet reduce performance on unseen data. We present \systemname{}, an automatic harness that turns a supplied model, target data, and a free-text request into a new-site adaptation workflow. Its LLM-driven \plannername{} uses model-site evidence to choose compatible experiments and revise subsequent adjustments. The harness requires evidence of improvement over the pretrained model, then delivers the selected model as a complete, runnable package. Evaluation spans 26 segmentation and classification transfers across architectures and compute settings, with segmentation annotation budgets ranging from zero-shot transfer to few-shot adaptation. With 20 local annotations, mean segmentation Dice rises from 0.483 to 0.774 across 10 transfers, a 60.1\% relative gain, with no harmful replacements. Mean classification AUROC rises from 0.810 to 0.837 across 13 transfers, a 3.4\% relative gain. Comparing adapted models with the pretrained model reduces harmful classification replacements by 30.8\% versus selecting the highest-scoring model on the available labels, using identical candidates. Seven public-model deployments demonstrate adaptation and reproducible package delivery across contrast and modality shifts. A CT-to-MRI study extends the evaluation to four LLM planners, including two open-weight models. We release the code on \href{https://github.com/Guhanxue/medagent}{GitHub}.
\end{abstract}

\section{Introduction}
Medical image analysis at a new hospital no longer has to begin with training a model from scratch. Public repositories now distribute trained model packages, including MONAI bundles for organ segmentation and TotalSegmentator for whole-body anatomical analysis~\citep{monai2025spleenbundle,monai2025swinunetrbundle,wasserthal2023totalsegmentator}. When one of these models addresses the same or a closely related task, a hospital can start from it rather than first collecting a large annotated cohort. Multi-center benchmarks and medical adaptation studies have documented substantial variation across hospitals, scanners, acquisition protocols, and patient populations~\citep{bandi2018camelyon,koh2021wilds,karani2021tta}. Source and target data can also differ in modality, annotation convention, image resolution, or preprocessing, changing the inputs, the expected outputs, or both~\citep{yu2025medsegtta}. Direct deployment can therefore lead to poor performance or complete failure~\citep{karani2021tta,xiong2026completeuda}.

When direct deployment fails, adaptation offers an alternative to rebuilding the model from scratch. Applying it requires choosing and configuring a method for the specific model-site pair. Domain adaptation, test-time adaptation, and few-shot repair offer adjustments such as input matching, normalization updates, probing, and fine-tuning~\citep{wang2021tent,niu2023towards,kumar2022lpft,lee2023surgical}. Their feasibility and effect depend on the supplied model and deployment setting. A BatchNorm-based method requires compatible normalization layers, while supervised fine-tuning requires labels and training compute; a method that helps one shift may harm another~\citep{zhao2023pitfalls,yu2025medsegtta}. With only a few annotations, an adapted model may score well on the labeled examples yet perform worse on unseen target data than the pretrained model~\mbox{\citep{cygert2024realistic,xiong2026completeuda}}. Applying these methods also requires a team to diagnose the mismatch, inspect the model and preprocessing, configure compatible experiments, interpret uncertain results, and package the selected model for inference. For clinical researchers without dedicated ML engineering support, these steps create a substantial barrier to model reuse. Automatic new-site deployment must therefore combine adaptation methods with model analysis, experiment planning, and controlled release.

Most existing automation focuses on model development rather than deployment at a new site. General and medical ML agents begin with task data and an optimization objective, then plan experiments, modify code, and train models~\citep{huang2023mlagentbench,feng2025m3builder,liu2026amid}. AutoMiSeg automatically assembles a zero-shot segmentation pipeline from foundation models and test-time adaptors~\citep{li2025automiseg}. \citet{xiong2026completeuda} study model selection within a precomputed pool of UDA models and validators. These systems automate model construction or candidate selection. New-site deployment starts from a trained model and requires analysis on local data, feasible adjustments under label and compute constraints, comparison with the pretrained model, and a runnable deployment package.

\begin{figure*}[t]
\centering
\includegraphics[width=\textwidth]{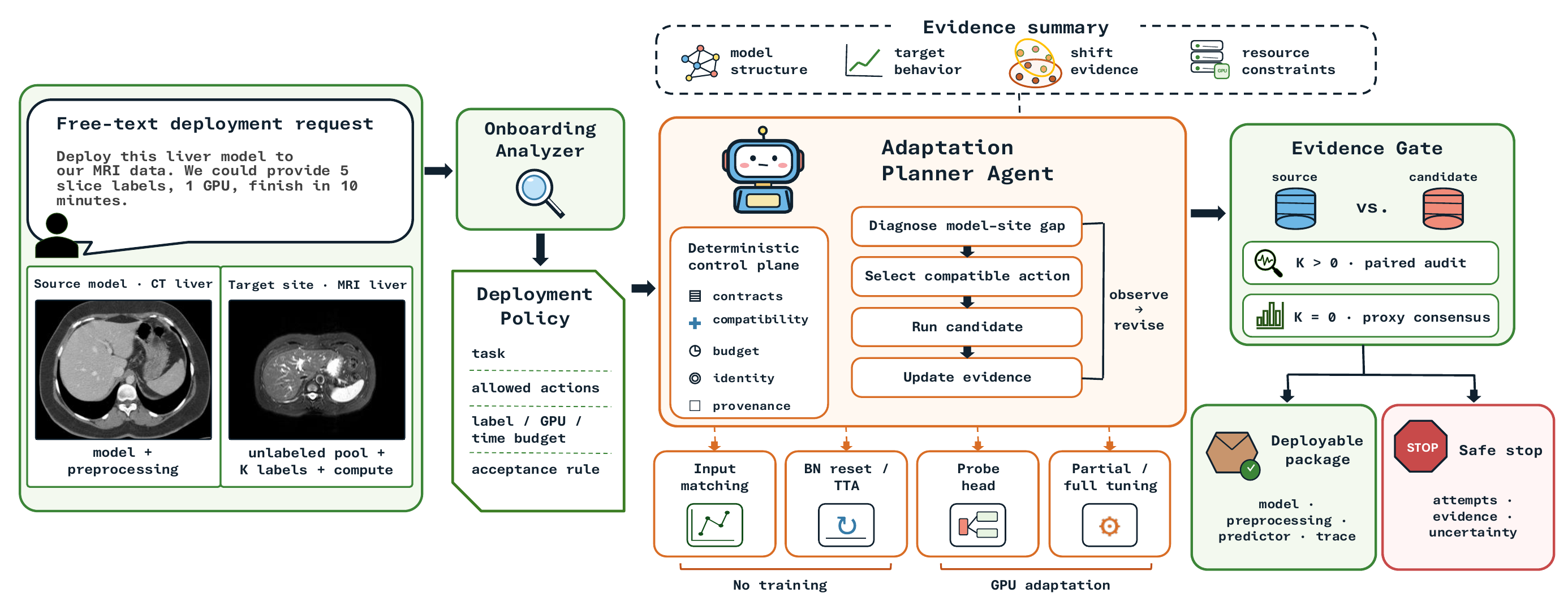}
\caption{\textbf{Automatic new-site model adaptation with the \harnessname{}.} A supplied model and target-hospital request enter the \onboardingname{}, which inspects the model-site pair and constructs the feasible action space. The LLM-driven \plannername{} iterates planning, tool execution, and evidence updates within deterministic compatibility and budget controls. The \gatename{} authorizes replacement. The harness then packages the pretrained model or an evidence-supported adapted model, and returns a safe-stop report when no tested option satisfies the deployment objective.}
\label{fig:overview}
\end{figure*}

We introduce the \harnessname{}, an automatic harness for post-training adaptation and new-site deployment, instantiated for segmentation and classification (Figure~\ref{fig:overview}). The system begins with a free-text deployment request, a supplied model, target-site data, and the available annotation and compute budgets. The \onboardingname{} translates these inputs into a validated deployment contract and inspects the model and target data. The \plannername{} uses this evidence to characterize the model-site gap, construct a compatible action space, and form an experiment policy for the current deployment. It then operates in a closed loop: select an adjustment, execute it, observe the result, and update the next action. This process allows the harness to move among input matching, normalization and test-time adaptation, probing, and fine-tuning as evidence accumulates.

Experiment planning and model release serve different roles in the harness. The \plannername{} chooses what to test, while deterministic controls enforce architectural compatibility, resource budgets, candidate identity, and provenance. The \gatename{} compares each candidate with the pretrained model using the target evidence available to the site. The final output is therefore a deployment package containing the pretrained or an evidence-supported adapted model. When the site specifies an absolute performance requirement and every tested option falls short, the harness returns a safe-stop report with the attempted adjustments and supporting evidence.

We evaluate segmentation and classification transfers across target sites, model architectures, annotation budgets, and GPU availability. Segmentation spans 0, 1, 5, and 20 labeled target cases. With 20 labels and supervised adaptation available, mean Dice rises from 0.483 to 0.774 across 10 transfers, with no harmful replacements. Across 13 classification transfers, mean AUROC rises from 0.810 to 0.837. Evidence-guided planning improves mean Dice from 0.596 for a fixed adjustment sequence to 0.629. Across four classification annotation draws, comparison with the pretrained model reduces harmful replacements by 30.8\% relative to choosing the model with the highest score on the small labeled selection set, using identical candidate models. All seven public-model deployments produce packages that reproduce the selected predictions exactly after reload.

\paragraph{Contributions.} Our contributions address three barriers to reusing medical-imaging models at new sites. \textbf{(1) Automating model reuse:} we introduce an end-to-end harness that turns a free-text request, an existing model, and target data into a model-specific analysis and adaptation workflow with a runnable deployment package. \textbf{(2) Adapting the strategy to the deployment:} we construct feasible experiments from model capabilities and local resources, then revise the adjustment policy using accumulated evidence. This closed-loop strategy improves on a fixed sequence of the same adjustment tools. \textbf{(3) Reducing harmful model replacements:} we separate candidate search from release and require evidence of improvement over the pretrained model. Controlled comparisons show that this verification reduces the risk of deploying an adapted model that scores well on a small labeled set yet performs worse on unseen target data than the pretrained model.

\section{Related Work}
\textbf{Test-time and few-shot adaptation.} BatchNorm (BN) statistic recalibration (BN reset)~\citep{schneider2020improving}, Tent~\citep{wang2021tent}, SAR~\citep{niu2023towards}, and EATA~\citep{niu2022efficient} are test-time adaptation (TTA) methods that adjust a pretrained (source) model using unlabeled data from the new site. Their behavior depends on batch size, initial model performance, normalization layers, and the change in data distribution~\citep{zhao2023pitfalls,karani2021tta}. A recent medical-segmentation benchmark finds that method rankings vary across modalities and that adaptation frequently degrades individual transfers~\citep{yu2025medsegtta}. Linear probing~\citep{kumar2022lpft}, surgical fine-tuning~\citep{lee2023surgical}, and weight interpolation~\citep{wortsman2022wiseft,wortsman2022soups} provide supervised alternatives. These methods form the harness action library; model inspection and resource constraints determine which are available for each deployment.

\textbf{Adaptation pipelines and model selection.} The selection rule can change TTA rankings, and selection using a small labeled set can outperform label-free model-selection rules~\citep{cygert2024realistic}. \citet{xiong2026completeuda} evaluate 13 label-free validators over a large pool generated by 10 unsupervised domain adaptation (UDA) algorithms. We adapt their small-label selection strategy to our classification benchmark, selecting by AUROC instead of accuracy. AutoMiSeg uses Bayesian optimization and a proxy validator to configure test-time adaptors for segmentation~\citep{li2025automiseg}. Our harness generates candidates from a supplied model under local resource constraints and compares each proposed replacement with the pretrained model.

\textbf{Agents for model development and maintenance.} General ML agents iteratively modify code and run experiments to optimize a task metric~\citep{huang2023mlagentbench}. M$^3$Builder automates data processing, environment configuration, debugging, and training, while AMID plans and verifies medical-imaging development pipelines~\citep{feng2025m3builder,liu2026amid}. The \harnessname{} addresses deployment after training: its planner selects registered adjustments to an existing model, and fixed procedures record the measurements supporting each model-selection decision. Such records support traceability of model updates~\citep{fda2023pccp}.

\section{Method: Automatic New-Site Model Adaptation}
\paragraph{Inputs and outputs.} The harness receives a pretrained model $f_s$, images from the new site $X_t$, a label budget $K$, trial and compute budgets, and a deployment objective. Images or statistics from the source domain and preprocessing metadata are optional. Each adjusted version is a candidate model. The default objective is improvement over the pretrained model; a request may also specify a minimum performance. The output is a complete deployment package or a report that the objective remains unmet, termed a safe stop.

\paragraph{Onboarding Analyzer.} The \onboardingname{} converts the request into a structured deployment contract specifying the model, data, budgets, and objective. Missing model or data references trigger a clarification; when annotation information is omitted, the harness uses $K{=}0$ and the conservative acceptance rule. The harness validates the contract and inspects the serialized model to identify its architecture family, task, outputs, normalization layers, and head. Data profiling supports NPZ arrays, per-case NPZ directories, and Decathlon-style NIfTI directories. These descriptions select the code interface for two-dimensional segmentation or classification. Three-dimensional tasks use registered loaders specifying intensity processing, spacing, orientation, and output-channel semantics; missing metadata produces a preprocessing error.

\paragraph{Action construction.} Initial predictions provide entropy, class frequencies or foreground area, and segmentation fragmentation. Image statistics are compared with source-domain references when available. The planner receives these measurements together with the model's capabilities and remaining resources. Compatibility rules restrict normalization methods to the relevant layers, probes to recognized output heads, and source-relative transforms to models with reference statistics. Supervised actions require target labels and training compute; three-dimensional actions additionally require a registered preprocessing contract.

\paragraph{Adaptation Planner.} At iteration $t$, the LLM receives a state $S_t=(c,\tau,E_t,C_t,B_t,F_t,\mathcal A_t)$ containing the deployment contract $c$, task descriptor $\tau$, accumulated evidence $E_t$, candidate table $C_t$, remaining budget $B_t$, action-validation feedback $F_t$, and permitted tools $\mathcal A_t$. The state and tool descriptions form the prompt. The LLM returns one structured action $a_t=(u_t,\theta_t)$, where $u_t\in\mathcal A_t$ specifies a diagnostic, adjustment, label request, or terminal decision. The arguments $\theta_t$ configure the action, including tuning capacity, learning rate, or training steps.

Executing the action updates the measurements, candidate records, and resource usage in $S_{t+1}$. The harness removes completed one-time diagnostics and rejects malformed, incompatible, or over-budget actions. All model changes pass through registered tools. The planner may propose a candidate model, retain the pretrained model, or stop; accepting an update requires approval from the \gatename{}.

\paragraph{Adjustment executors.} The action library includes input matching, normalization adaptation, output-layer training on frozen features, low-rank adaptation (LoRA), and partial fine-tuning (Table~\ref{tab:actions}). Each successful experiment records a candidate model and its measurements for the planner.

\begin{table}[t]
\centering
\footnotesize
\caption{Adjustment families exposed to the planner after capability and resource filtering. The GPU requirement is mixed for normalization/TTA because the family contains both forward-only BN reset and gradient-based methods.}
\label{tab:actions}
\begin{tabular}{@{}lp{3.7cm}ccp{3.7cm}@{}}
\toprule
family & examples & labels & GPU? & principal requirement \\
\midrule
input matching & histogram, stain, frequency & 0 & no & compatible input semantics \\
normalization/TTA & BN reset, Tent, SAR, EATA & 0 & mixed & supported trainable statistics \\
frozen probe & linear or per-pixel head & $\geq 5$ & yes & recognized output head \\
model tuning & head, partial, LoRA, surgical & $\geq 5$ & yes & training compute \\
\bottomrule
\end{tabular}
\end{table}

\paragraph{Evidence Gate.} Without target labels, the gate checks prediction confidence and plausibility relative to the pretrained model; segmentation also uses shape changes and visual comparison. Candidates must pass structural checks and a combined score threshold. With labels, the pretrained and candidate models are compared on the same labeled selection set. Leave-one-out or held-out evaluation excludes each scored example from the corresponding model fit. Paired resampling estimates variation in per-case gains; a minimum score improvement is required when paired predictions are unavailable. Search and final selection reuse the selection set; resampling describes within-set variation for the selected predictions. Appendix~\ref{sec:gate_details} specifies the statistics and thresholds.

\paragraph{Deployment packaging.} A deterministic packager exports the selected model with preprocessing, inference code, environment metadata, and a report linking the selected model to its supporting measurements.

\section{Experimental Setup}
\paragraph{Benchmarks.} The main benchmark comprises 26 model-site transfers: 13 two-dimensional segmentation and 13 classification pairs. Each architecture--target pair counts as one transfer. The segmentation budget study uses ten U-Net transfers, while the planning comparison uses a separate eleven-transfer suite. We evaluate annotation and compute budgets, experiment planning, model-selection rules, and deployment from public model packages.

\paragraph{Tasks and models.} Segmentation spans six anatomies: prostate using RUNMC$\rightarrow$HK~\citep{bloch2015nci,litjens2014promise12}; fundus using RIGA+ and REFUGE$\rightarrow$RIM-ONE DL~\citep{almazroa2018riga,orlando2020refuge,fumero2020rimone}; cardiac using M\&Ms~\citep{campello2021mms}; abdominal using CHAOS and AMOS~\citep{kavur2021chaos,ji2022amos}; whole-heart using MMWHS~\citep{zhuang2019mmwhs}; and brain using BraTS~\citep{menze2015brats}. The primary budget study fixes the architecture family to U-Net across 10 transfers. A controlled comparison varies U-Net, SegFormer, nnU-Net, and STU-Net on the same prostate shift across $K\in\{0,1,5,20\}$, while a broader stress test measures normalization adaptation across SegFormer, SwinUNETR, SegResNet, and three-dimensional U-Nets. Classification covers diabetic retinopathy, dermatology, histopathology~\citep{koh2021wilds,bandi2018camelyon}, chest radiography, and breast histology.

\paragraph{Annotation and compute regimes.} Segmentation uses a fixed target test set and label budgets $K\in\{0,1,5,20\}$. At $K{=}0$, the harness constructs compatible BN-reset, Tent, SAR, and EATA candidates~\citep{schneider2020improving,wang2021tent,niu2023towards,niu2022efficient} and applies its zero-label selection rules. At $K{=}1$, the harness uses the single annotation to rank prebuilt candidates adapted without labels. Budgets $K{=}5$ and $K{=}20$ unlock probing and fine-tuning. Models trained with target labels use out-of-fold or held-out evaluation; test labels remain withheld until model selection is complete. A compute ablation disables all supervised training at $K{=}5$ and $K{=}20$. Classification uses 12 target labels, except 18 for nine-class CRC-VAL (two per class).
\paragraph{Annotation units and data splits.} In the two-dimensional segmentation study, $K$ counts labeled images or slices. A fixed twenty-image adaptation pool is shared across budgets, and the remaining target images form the test set. Benchmark masks identify foreground-containing images for the pool; $K$ counts the annotations subsequently revealed to the controller, excluding this pool-construction effort. Images are assigned individually, allowing slices from one patient to occur in both partitions. Classification budgets count images or patches; patient grouping is undocumented in the available split records. The external three-dimensional study counts labeled scans and separates adaptation and test scan identifiers.

\paragraph{LLM implementation.} We evaluate GPT-5.2, GPT-5.6 Luna, Qwen3-VL-4B-Instruct, and Qwen3-VL-8B-Instruct as action planners. The broader benchmarks use \texttt{gpt-5.2-2025-12-11} through Azure OpenAI with temperature zero and a 2,000-token response limit; the four-planner CT-to-MRI comparison is detailed in Appendix~\ref{sec:planner_comparison}. The policy receives text prompts and emits one tool action per step. For renderable segmentation candidates, GPT-5.2 compares pretrained and adapted model overlays on up to six target images. The $K{=}0$ consensus combines this pairwise preference with prediction-derived signals. Target-test labels never enter planner prompts.

\paragraph{Baselines and controls.} Baselines include direct deployment, fixed BN reset, Tent, SAR, EATA, fixed fine-tuning, and a fixed-sequence baseline (scripted ladder) with the same adjustment library and Evidence Gate. Segmentation partial fine-tuning updates the final 40\% of parameterized leaf modules and the output head using Adam, learning rate $10^{-3}$, and 80 steps. A label split supports early stopping at $K{=}20$; smaller budgets use all $K$ labels. Classification baselines update the head or the full model using AdamW for 80 steps. The planning comparison fixes the available tools and acceptance rules; compute use depends on the experiments each policy chooses.

Selection baselines choose the model with minimum prediction entropy at $K{=}0$ or maximum AUROC on the labeled selection set for classification~\citep{xiong2026completeuda}. The classification comparison gives both rules identical candidates and labels over four annotation draws. Registration order breaks ties, with the pretrained model registered first. Source-label-dependent validators are excluded. After benchmark pool construction, test labels are withheld from training, planner prompts, and model selection.

\paragraph{Outcome measures.} We report target-test Dice or AUROC and harmful replacements, defined as updates that reduce test performance relative to the pretrained model. For $K{=}0$, coverage is the fraction of transfers receiving an adapted model. Decision value is mean test performance across transfers, using the pretrained-model score when the harness stops without an update.

\paragraph{Annotation draws and integration tests.} Classification uses four seeded annotation draws; standard deviations summarize the four draw-level means. Each segmentation budget setting and planning comparison uses one run. Two CHAOS experiments evaluate candidate selection and end-to-end deployment: a standardized comparison of seven candidates and adaptation starting from a free-text request. Both use five annotations. The end-to-end experiment evaluates the exported predictor on 401 images outside its labeled selection set and checks prediction equality after two independent reloads. The two evaluation settings are described in Appendix~\ref{sec:chaos_trace}.

\paragraph{External public models.} Seven deployment configurations use three MONAI bundles: MSD-spleen U-Net, BTCV SwinUNETR, and TotalSegmentator SegResNet~\citep{monai2025wholebodybundle}. Targets include non-contrast CT, MSD spleen CT, and AMOS MRI~\citep{ji2022amos}. Registered loaders preserve preprocessing and extract the spleen output. Each deployment uses one GPU and a target cohort of 30 three-dimensional scans. With five labeled target scans, three are used to train candidate models and two to select among them. With twenty labeled scans, thirteen are used for training and seven for selection. Final performance is measured on ten separate scans whose labels are withheld from adaptation and model selection; these test scans are unchanged across annotation budgets. Search ends at the trial limit without improvement, and the controller selects among the tested models. Exported models are reloaded to check prediction equality on the test scans. Deployment outcomes are summarized in Appendix~\ref{sec:public_attempts}.

\section{Results}
\subsection{Adaptation changes with evidence, compute, and model}
\begin{figure}[t]
\centering
\includegraphics[width=\textwidth]{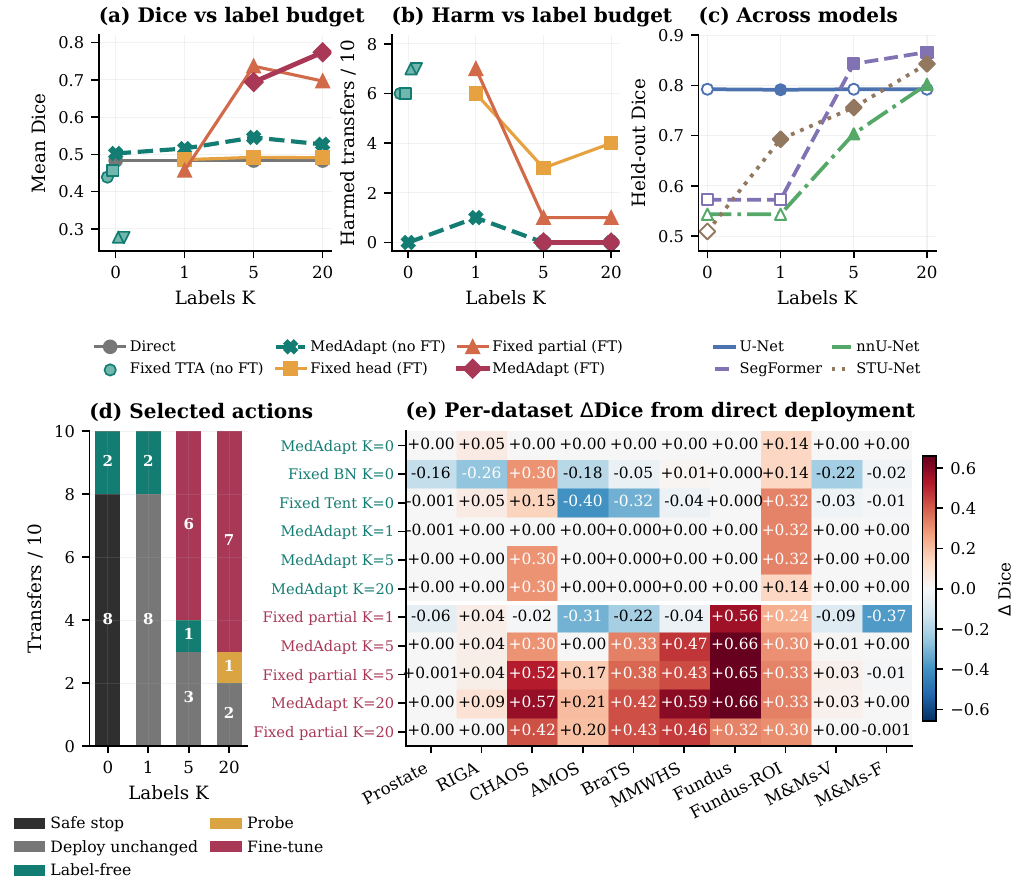}
\caption{\textbf{Segmentation adaptation across annotation and compute budgets.} (a,b) Mean Dice and harmful replacements across 10 U-Net transfers. Teal denotes no supervised target training, warm colors denote training availability, and gray denotes direct deployment. The offset $K{=}0$ markers show BN reset, Tent, SAR, and EATA, from left to right. (c) Target-test Dice across four models on the same RUNMC-to-HK prostate shift. Open markers show pretrained-model performance, including cases where the harness stops without an update; filled markers deploy an adapted model. (d) Selected actions. (e) Per-dataset changes from direct deployment; teal and warm row labels indicate supervised training unavailable and available, respectively. All test scores are measured after model selection is complete.}
\label{fig:budget}
\end{figure}

\input{table_seg_budget}
With no target labels, accepting only selected updates improves average performance. Each fixed TTA method degrades a majority of the ten transfers, whereas the harness accepts two beneficial updates and stops on the remaining eight (Table~\ref{tab:seg_budget}). Mean decision value increases from 0.483 to 0.502. The accepted updates are on RIGA and Fundus-ROI.

Prediction confidence alone selects poorly among these candidate models. Minimum-entropy selection replaces nine pretrained models and degrades five transfers, reducing mean performance below direct deployment (Appendix Table~\ref{tab:k0_selector}). The prediction-based score and visual model comparison also select harmful updates when used individually. Requiring agreement among these signals accepts updates on two transfers, both of which improve.

With one annotation, candidate selection produces fewer harmful updates than fixed partial fine-tuning. The harness reaches 0.516 mean Dice, while partial fine-tuning degrades seven transfers and falls below direct deployment. On Fundus, training on one image improves the pretrained model from zero Dice to 0.560 Dice. The harness retains the pretrained model because its one-label policy evaluates models adapted without labels. Using the label for selection reduces harmful updates across the suite, while leaving this training-based improvement unrealized.

Supervised adaptation becomes more effective with larger budgets. At five labels, fixed partial fine-tuning has higher mean Dice than the harness, with one harmful transfer. At twenty labels, the harness leads, improving eight transfers and retaining two pretrained models for a mean of 0.774, compared with 0.697 for fixed partial fine-tuning. Its choices include a probe for Fundus, LoRA for the M\&Ms vendor shift, and partial fine-tuning for several other sites.

The Fundus recovery illustrates why the search includes adjustments with different capacities. Training a new output layer on frozen features raises the full-field model from zero Dice to 0.657, showing that its learned features remain useful. The model using region-of-interest preprocessing benefits from partial fine-tuning (Appendix Table~\ref{tab:seg_tasks}). These two pretrained models face the same disc-centered target, yet their selected adjustments differ. Initial model performance and preprocessing history therefore matter alongside the target dataset.

Training compute determines how much of the larger label budget can be used for recovery. With twenty labels, disabling supervised adaptation limits mean Dice to 0.527; enabling it raises the mean to 0.774 (Figure~\ref{fig:budget}(a,b)). Using additional labels only to rank models adapted without labels has a smaller effect: twenty-label mean Dice remains below the five-label result.

Model architecture changes which adjustment helps on the same target dataset. On prostate, the relatively strong pretrained U-Net is retained across budgets, while SegFormer and nnU-Net benefit once supervised adaptation becomes available (Figure~\ref{fig:budget}(c)). STU-Net improves earlier through SAR selected with one label, then benefits further from partial fine-tuning.

\paragraph{Compatibility and adaptation benefit.} BN-based tools leave the LayerNorm-, InstanceNorm-, and GroupNorm-only models unchanged (Appendix Table~\ref{tab:arch}). For SegFormer~\citep{xie2021segformer}, a BatchNorm output head makes both BN reset and Tent active. Reset produces a small gain; Tent causes a large loss. The BatchNorm-based three-dimensional U-Nets likewise respond differently to the same methods. Architecture inspection rules out adjustments that leave the model unchanged; local evaluation determines which remaining adjustments help.

\paragraph{Classification.} Across the classification transfers, the harness improves mean AUROC and reduces the frequency of harmful updates relative to fixed fine-tuning. Head fine-tuning achieves higher mean AUROC, illustrating a trade-off between average performance and the frequency of degradation. Detailed results, model-specific outcomes, and variation across annotation draws are reported in Appendix~\ref{sec:classification_results}.

\paragraph{End-to-end adaptation.} The CHAOS end-to-end experiment starts from a free-text request and exports a model selected with five labeled images. Partial fine-tuning improves Dice from 0.117 to 0.256 on the 401 remaining images, and two independent reloads reproduce all twenty checked masks. This agreement confirms that packaging preserves the selected model's predictions. Appendix~\ref{sec:chaos_trace} reports the separate standardized search.

\subsection{Adaptive search improves on a fixed sequence}
The planner achieves 0.629 mean Dice on the eleven-transfer, five-label suite, compared with 0.596 for the scripted ladder. Both policies use the same adjustment library and acceptance rules. The ladder follows a preset order. The planner configures each experiment using measurements from the model and local data and the results of previous experiments.

Both policies improve nine transfers and preserve pretrained-model performance on the remaining two. The planner's advantage therefore comes from the magnitude of recovery across the suite. Even with the same available adjustments, the choice and configuration of experiments affect performance. The planner responds to observed local results, while both policies use the same rules to decide whether to retain the pretrained model. 

The CT-to-MRI comparison extends this workflow to four LLM planners. GPT-5.2, GPT-5.6 Luna, and both Qwen planners produce adapted models and reproduce all ten test masks after reload. Their searches select different updates and training configurations, yielding different test Dice under the same adaptation and selection rules. This supports using both proprietary and open-weight planners within the harness (Appendix~\ref{sec:planner_comparison}). 

\subsection{Comparison with the pretrained model reduces harmful updates}
\begin{table}[t]\centering\footnotesize
\caption{Classification selection with 12 labels (18 for nine-class CRC-VAL): 13 transfers over four draws yield 52 decisions. AUROC is mean $\pm$ population standard deviation of the four draw-level means. Both rules use identical candidates and labels; the score-based baseline selects the model with the highest AUROC on the labeled selection set~\citep{xiong2026completeuda}. Bold marks the best values.}\label{tab:verifier}
\begin{tabular}{lccc}\toprule
replacement rule & mean AUROC $\uparrow$ & harmed $\downarrow$ & worst $\Delta$ $\uparrow$ \\ \midrule
highest selection AUROC & $0.827\pm0.015$ & 13/52 & $-0.343$ \\
\systemname{} & $\boldsymbol{0.839\pm0.006}$ & \textbf{9/52} & $\boldsymbol{-0.184}$ \\ \bottomrule
\end{tabular}\end{table}
Comparing each proposed update with the pretrained model improves selection among the same candidate models. Across 52 classification decisions, this rule reduces harmful replacements from thirteen to nine and improves mean test AUROC relative to choosing the highest score on the labeled selection set (Table~\ref{tab:verifier}). The worst-case loss also decreases. Because candidates and labels are identical, these differences arise at the selection step.

The pretrained-model comparison also reduces variation across annotation draws: the standard deviation of draw-level mean AUROC falls from 0.015 to 0.006. This aggregate stability accompanies frequent changes in the selected action across draws. Different small selection sets can favor different adjustments while producing similar average test performance. Comparison with the pretrained model provides a common reference for these adjustments and complements the planner's search.

An absolute performance requirement can reject an update that improves on a weak pretrained model. In the separate prostate-classification stress test, Tent raises target AUROC from 0.479 to 0.508, the best result among the tested adjustments (Appendix Table~\ref{tab:safe_stop}). A requested floor above 0.508 produces a stop. The criterion prevents relative improvement alone from authorizing deployment when all tested models remain below the site's requirement.

\subsection{Deployment from external public models}
\input{table_external_public}
All seven public-model configurations yield verified deployment packages: five adapted predictors and two unchanged pretrained models (Table~\ref{tab:external_public}). Each package reproduces all ten test masks after reload. Completed executions take 183--1,039 seconds on one GPU, including prediction checks after reload.

On CT, the value of adaptation depends on the starting model and the evidence for replacement. SegResNet already achieves 0.951 Dice on MSD spleen CT, and its scores on the selection set favor retaining the pretrained model over the tested updates. On non-contrast CT, the harness selects LoRA for SwinUNETR and retains the pretrained U-Net. The adapted U-Net improves substantially on the test set, while its gain of 0.028 on the selection set falls below the required improvement margin. Retention therefore has different consequences: it preserves an accurate SegResNet model and misses a beneficial U-Net update.

The twenty-label MRI configurations outperform their five-label counterparts for both model families. U-Net moves from normalization adaptation to learned LoRA updates, while SwinUNETR uses partial fine-tuning at both budgets. This pattern connects annotation availability to the kind of recovery the model can support: normalization adjusts feature statistics, whereas supervised tuning learns from the target appearance and anatomy. MRI performance remains below the corresponding non-contrast CT results, making modality transfer the more demanding setting in these deployments.

\section{Discussion}
Model reuse requires choosing an adjustment and evaluating its benefit. The experiments connect model inspection to compatible adjustments, adaptive search to larger gains, and pretrained-model comparison to fewer harmful updates. The architecture comparisons, budget study, and matched-candidate analysis support these complementary roles. Each selected model is linked to its experiment record, acceptance decision, and prediction checks after reload.

Annotations serve two purposes: training an adapted model and evaluating its performance. Reserving a single label for selection reduces tuning failures across the suite and leaves the Fundus model unchanged. At larger budgets, training becomes essential to the strongest segmentation gains. For a new deployment, annotation planning should therefore account for the adjustments that the model and compute budget can support. Improvement requires both an effective adapted model and an evaluation that identifies its benefit.

Acceptance criteria determine which errors arise during model selection. Requiring a minimum improvement can reject a beneficial update when the selection set understates its gain, as in the public U-Net example. The same rule can accept harmful updates when the selection set overstates the gain. Frequency and severity of harm must therefore be evaluated alongside mean performance. An absolute performance requirement serves a separate purpose: it rules out deployment when even the best tested model falls below the requested minimum performance.

\paragraph{Limitations.} The four-planner comparison covers one CT-to-MRI transfer; the broader benchmarks use GPT-5.2. Segmentation and planning comparisons use single runs; the fixed-sequence baseline uses compute according to its chosen experiments. Acceptance thresholds are heuristic, search and final selection reuse the labeled evaluation set, and the classification comparison measures gains in true-class probability. The two-dimensional results describe image-level transfer. Clinical translation requires independent patient-level local validation against the intended performance requirements.

\section{Conclusion}
\systemname{} turns a pretrained medical-imaging model and a local adaptation request into a complete, runnable package. Across the evaluated transfers, adaptive planning improves on a fixed experiment sequence, comparison with the pretrained model reduces harmful updates among identical candidates, and supervised training produces the largest segmentation gains. Each package records the experiments and measurements supporting its selected model.

\clearpage
\bibliographystyle{plainnat}
\bibliography{references}

\appendix
\input{appendix_classification}
\input{appendix_supporting_tables}

\section{Evidence Gate details}\label{sec:gate_details}
\paragraph{Thresholds.} The coefficients, margins, and percentiles are manually specified. Structural checks exclude implausible predictions, gain floors require a minimum observed improvement, and bootstrap percentiles set the evidence requirement for each objective. These values serve as heuristic decision settings; using them to bound the risk of a harmful update requires independent calibration.

\paragraph{Classification statistic.} The gate resamples the per-case gain $p_c(y_i\mid x_i)-p_s(y_i\mid x_i)$ in probability assigned to the true class. AUROC measures ranking across cases. A candidate can increase true-class probability while decreasing AUROC. Final classification outcomes are evaluated using target-test AUROC.

\paragraph{Reuse of selection labels.} Out-of-fold and holdout predictions separate fitting from scoring for each candidate. The labeled evaluation set is reused across experiments and final model selection. The bootstrap quantifies within-set variation conditional on these adaptively selected predictions.

At $K{=}0$, the gate compares segmentation candidates with the pretrained model using a predefined consensus of prediction-derived signals. For statistics $s=(H,r,n)$, representing mean entropy, foreground fraction, and component count, the scalar proxy is
\[
q(s)=0.50\max(0,1-H)+0.35\!\left[1-\min\!\left(1,\frac{|r-r_0|}{4\max(r_0,0.05)}\right)\right]
+0.15\,[1+\max(0,n-2)]^{-1},
\]
where $r_0$ is the expected foreground fraction. Let $\Delta q=q(s)-q(s_s)$, $\rho_r=r/r_s$, and $\rho_n=n/n_s$, clipping fractions at $10^{-6}$ and component counts at 1 before forming ratios. When segmentation overlays can be rendered, a vision-language model (VLM) compares candidate and pretrained-model predictions, giving the fraction of comparisons favoring the candidate, $p$. Preprocessing adjustments that do not support overlay rendering omit this term. These signals define the zero-label selection score:
\[
z=\Delta q+\begin{cases}0.20(p-0.5),&p\text{ available},\\-0.05,&\text{otherwise},\end{cases}
-0.02\min(2,|\log\rho_r|)-0.01\min(2,|\log\rho_n|).
\]
A candidate is eligible when $\Delta q\geq0$, $z\geq0.03$, and structural checks reject implausibly small or large foreground regions ($r\notin[0.2r_0,4r_0]$), empty masks, and severe fragmentation. A characterized semantic shift additionally requires $p\geq0.67$. The eligible candidate with the largest $z$ is proposed to the gate. Updates accepted without labels are marked as based on estimated performance.

Zero-label classification checks class collapse and requires a confidence-based quality gain of at least 0.03 over the pretrained model.

With labels, fine-tuned segmentation candidates use leave-one-out refits at $K{=}5$ and held-out evaluation at $K{=}20$. Classification fine-tuning uses a holdout, while probes use out-of-fold predictions. Candidates adapted without labels are evaluated using all available labels. Paired verification compares per-case Dice gains for segmentation and true-class probability gains for classification using 2,000 bootstrap resamples. The lower percentile must exceed zero: the 10th percentile for the conservative objective and the 25th for maximum performance. A mean-gain floor of 0.005 also applies, except for classification probes. When paired evidence is unavailable, including classifier fine-tuning with fewer than six held-out cases, replacement requires a score improvement of 0.05 on the labeled evaluation set. Public three-dimensional runs apply the same 0.05 margin to their held-out selection scans.

\section{CHAOS adaptation and deployment}\label{sec:chaos_trace}
The CHAOS experiments evaluate adaptation in two settings with distinct evaluation partitions. The standardized comparison uses five annotations to select among seven candidates. The search selects BN reset, improving Dice from 0.120 to 0.391.

The end-to-end experiment begins with the same CT-trained model and a free-text request. With five labeled selection images, the harness selects partial fine-tuning and improves Dice from 0.117 to 0.256 on the remaining 401 images. Two independent reloads reproduce all twenty checked masks, confirming that the delivered package preserves the selected predictions.

\input{appendix_public_attempts}

\section{Per-transfer outcomes}
Table~\ref{tab:seg_tasks} records the $K{=}20$ segmentation decisions, including the controlled four-model comparison; Table~\ref{tab:cls_tasks} records the classification decisions. Each architecture evaluated on the same target constitutes a separate transfer. Shift labels summarize the dominant observed mismatch.

\begin{table}[H]\centering\scriptsize
\caption{Per-transfer segmentation Dice at $K{=}20$. Fundus rows distinguish full-field-of-view (FOV) and region-of-interest (ROI) preprocessing; the RIM-ONE target is disc-centered in both cases.}\label{tab:seg_tasks}
\begin{tabular}{lllrrl}\toprule
transfer & architecture & dominant shift & source & output & selected action \\ \midrule
AMOS CT$\rightarrow$MRI multiorgan & U-Net & modality & 0.723 & 0.928 & partial fine-tuning \\
BraTS T1$\rightarrow$FLAIR tumor & U-Net & sequence & 0.449 & 0.871 & partial fine-tuning \\
CHAOS CT$\rightarrow$MRI liver & U-Net & modality & 0.089 & 0.662 & partial fine-tuning \\
REFUGE full-FOV$\rightarrow$RIM-ONE & U-Net & scale/preproc. & 0.000 & 0.657 & final-layer probe \\
REFUGE ROI$\rightarrow$RIM-ONE & U-Net & intensity/scale & 0.383 & 0.715 & partial fine-tuning \\
MMWHS CT whole heart & U-Net & contrast & 0.039 & 0.633 & partial fine-tuning \\
M\&Ms Siemens$\rightarrow$GE & U-Net & intensity & 0.773 & 0.801 & LoRA fine-tuning \\
M\&Ms 1.5T$\rightarrow$3T & U-Net & mixed & 0.842 & 0.842 & retain source model \\
Prostate RUNMC$\rightarrow$HK & U-Net & intensity & 0.792 & 0.792 & retain source model \\
Prostate RUNMC$\rightarrow$HK & SegFormer & intensity & 0.573 & 0.866 & partial fine-tuning \\
Prostate RUNMC$\rightarrow$HK & nnU-Net & intensity & 0.544 & 0.802 & partial fine-tuning \\
Prostate RUNMC$\rightarrow$HK & STU-Net & intensity & 0.509 & 0.843 & partial fine-tuning \\
RIGA BinRushed$\rightarrow$Magrabia & U-Net & mixed & 0.744 & 0.838 & partial fine-tuning \\ \bottomrule
\end{tabular}
\end{table}

\begin{table}[H]\centering\scriptsize
\caption{Per-transfer classification AUROC for the annotation draw summarized in Table~\ref{tab:cls}, using 12 labels (18 for nine-class CRC-VAL).}\label{tab:cls_tasks}
\begin{tabular}{lllrrl}\toprule
transfer & architecture & diagnosed shift & source & output & action \\ \midrule
APTOS$\rightarrow$DDR & ResNet18 & mixed & 0.827 & 0.830 & head fine-tuning \\
APTOS$\rightarrow$EyePACS & ResNet18 & mixed & 0.716 & 0.704 & head fine-tuning \\
APTOS$\rightarrow$Messidor-2 & ResNet18 & mixed & 0.730 & 0.623 & partial fine-tuning \\
BreakHis 40$\times\rightarrow$400$\times$ & ResNet18 & resolution & 0.801 & 0.866 & head fine-tuning \\
CAMELYON17 cross-hospital & ResNet18 & intensity & 0.931 & 0.931 & retain source model \\
CAMELYON17 cross-hospital & Swin-Tiny & intensity & 0.981 & 0.997 & source histogram match \\
CAMELYON17 cross-hospital & ViT-small & intensity & 0.703 & 0.911 & head fine-tuning \\
NCT-CRC$\rightarrow$Chaoyang & ResNet18 & mixed & 0.727 & 0.872 & head fine-tuning \\
EyePACS$\rightarrow$APTOS & ResNet18 & mixed & 0.963 & 0.967 & head fine-tuning \\
HAM10000$\rightarrow$ISIC & ResNet18 & contrast & 0.767 & 0.767 & retain source model \\
HAM10000$\rightarrow$PAD-UFES & ResNet18 & modality & 0.608 & 0.643 & LoRA fine-tuning \\
NIH$\rightarrow$CheXpert & ResNet18 & contrast & 0.788 & 0.786 & head fine-tuning \\
NCT-CRC$\rightarrow$CRC-VAL & ResNet18 & contrast & 0.988 & 0.988 & retain source model \\ \bottomrule
\end{tabular}
\end{table}

\end{document}

%% file: table_seg_budget.tex
\begin{table}[t]
\centering
\scriptsize
\setlength{\tabcolsep}{1.7pt}
\renewcommand{\arraystretch}{1.02}
\caption{Per-dataset segmentation Dice across target-label budgets using the same U-Net architecture. Fixed rows apply one adjustment recipe to every transfer; \systemname{} analyzes each model-site pair and selects among compatible actions. Mean Dice uses the pretrained-model score when the harness stops without an update. Best values within each annotation-budget block are bold. M\&Ms-V/F denote vendor and field-strength shifts.}
\label{tab:seg_budget}
\resizebox{\textwidth}{!}{%
\begin{tabular}{lcccccccccccc}
\toprule
Policy & Prostate & RIGA & CHAOS & AMOS & BraTS & MMWHS & Fundus & Fundus-ROI & M\&Ms-V & M\&Ms-F & Mean $\uparrow$ & Harm $\downarrow$ \\
\midrule
Direct deployment & 0.792 & 0.744 & 0.089 & 0.723 & 0.449 & 0.039 & 0.000 & 0.383 & 0.773 & 0.842 & 0.483 & 0 \\
\midrule
Fixed BN reset, $K{=}0$ & 0.635 & 0.483 & \textbf{0.385} & 0.539 & 0.404 & 0.051 & 0.000 & 0.520 & 0.549 & 0.822 & 0.439 & 6 \\
Fixed Tent, $K{=}0$ & 0.791 & \textbf{0.796} & 0.237 & 0.324 & 0.131 & 0.000 & 0.000 & \textbf{0.707} & 0.746 & 0.835 & 0.457 & 6 \\
Fixed SAR, $K{=}0$ & 0.755 & 0.000 & 0.224 & 0.490 & 0.000 & \textbf{0.113} & 0.000 & 0.000 & 0.506 & 0.691 & 0.278 & 7 \\
Fixed EATA, $K{=}0$ & 0.756 & 0.000 & 0.223 & 0.490 & 0.000 & 0.111 & 0.000 & 0.000 & 0.497 & 0.688 & 0.277 & 7 \\
\systemname{}, $K{=}0$ & \textbf{0.792} & \textbf{0.796} & 0.089 & \textbf{0.723} & \textbf{0.449} & 0.039 & 0.000 & 0.520 & \textbf{0.773} & \textbf{0.842} & \textbf{0.502} & \textbf{0} \\
\midrule
Fixed head, $K{=}1$ & \textbf{0.795} & 0.734 & 0.085 & 0.698 & 0.448 & 0.015 & 0.000 & 0.444 & \textbf{0.812} & 0.827 & 0.486 & 6 \\
Fixed partial, $K{=}1$ & 0.729 & \textbf{0.785} & 0.073 & 0.411 & 0.230 & 0.003 & \textbf{0.560} & 0.623 & 0.688 & 0.468 & 0.457 & 7 \\
\systemname{}, $K{=}1$ & 0.791 & 0.744 & \textbf{0.089} & \textbf{0.723} & \textbf{0.449} & \textbf{0.039} & 0.000 & \textbf{0.707} & 0.773 & \textbf{0.842} & \textbf{0.516} & \textbf{1} \\
\midrule
Fixed head, $K{=}5$ & \textbf{0.793} & 0.743 & 0.089 & 0.735 & 0.448 & 0.051 & 0.000 & 0.430 & 0.789 & \textbf{0.842} & 0.492 & 3 \\
Fixed partial, $K{=}5$ & \textbf{0.793} & 0.781 & \textbf{0.604} & \textbf{0.892} & \textbf{0.826} & 0.469 & 0.650 & \textbf{0.709} & \textbf{0.807} & 0.836 & \textbf{0.737} & 1 \\
\systemname{}, $K{=}5$ & 0.792 & \textbf{0.785} & 0.385 & 0.723 & 0.780 & \textbf{0.508} & \textbf{0.658} & 0.685 & 0.782 & \textbf{0.842} & 0.694 & \textbf{0} \\
\midrule
Fixed head, $K{=}20$ & \textbf{0.792} & 0.744 & 0.089 & 0.736 & 0.448 & 0.054 & 0.000 & 0.426 & 0.779 & \textbf{0.843} & 0.491 & 4 \\
Fixed partial, $K{=}20$ & \textbf{0.792} & 0.744 & 0.508 & 0.921 & \textbf{0.874} & 0.501 & 0.325 & 0.687 & 0.773 & 0.841 & 0.697 & 1 \\
\systemname{}, $K{=}20$ & \textbf{0.792} & \textbf{0.838} & \textbf{0.662} & \textbf{0.928} & 0.871 & \textbf{0.633} & \textbf{0.657} & \textbf{0.715} & \textbf{0.801} & 0.842 & \textbf{0.774} & \textbf{0} \\
\bottomrule
\end{tabular}%
}
\end{table}

%% file: table_external_public.tex
\begin{table}[t]
\centering
\footnotesize
\caption{\textbf{Verified deployments from seven public-model configurations.} U-Net v0.6.1~\citep{monai2025spleenbundle}, SwinUNETR v0.5.8~\citep{monai2025swinunetrbundle}, and SegResNet v0.2.7 are evaluated by Dice on 10 held-out scans per target. Direct and delivered scores refer to the pretrained and packaged models. The diagnosis combines the inferred data shift and normalization layers: BN, BatchNorm; IN/LN, InstanceNorm/LayerNorm; GN, GroupNorm. CE/NC denote contrast-enhanced/non-contrast CT; TotalSeg denotes TotalSegmentator. Each package reproduces all 10 masks after reload.}
\label{tab:external_public}
\setlength{\tabcolsep}{3pt}
\begin{tabular}{@{}lllcrrl@{}}
\toprule
model & source $\rightarrow$ target & model/site diagnosis & $K$ & direct $\uparrow$ & delivered $\uparrow$ & outcome \\
\midrule
U-Net & MSD CE $\rightarrow$ TotalSeg NC & mixed; BN & 5 & 0.386 & 0.386 & retain source \\
SwinUNETR & BTCV CE $\rightarrow$ TotalSeg NC & mixed; IN/LN & 5 & 0.669 & \textbf{0.705} & LoRA \\
SegResNet & TotalSeg CT $\rightarrow$ MSD CT & intensity; GN & 5 & 0.951 & 0.951 & retain source \\
\midrule
U-Net & MSD CT $\rightarrow$ AMOS MRI & modality; BN & 5 & 0.000 & \textbf{0.120} & Tent \\
SwinUNETR & BTCV CT $\rightarrow$ AMOS MRI & modality; IN/LN & 5 & 0.040 & \textbf{0.191} & partial fine-tuning \\
U-Net & MSD CT $\rightarrow$ AMOS MRI & modality; BN & 20 & 0.000 & \textbf{0.233} & LoRA \\
SwinUNETR & BTCV CT $\rightarrow$ AMOS MRI & modality; IN/LN & 20 & 0.040 & \textbf{0.264} & partial fine-tuning \\
\bottomrule
\end{tabular}
\end{table}

%% file: appendix_classification.tex
\section{Classification results}\label{sec:classification_results}

\input{table_classification}
The harness improves mean AUROC over the pretrained models and reduces the number of degraded transfers relative to fixed fine-tuning (Table~\ref{tab:cls}). Head fine-tuning yields higher mean AUROC and a smaller worst-case loss than the harness. Its harmful updates span six transfers, compared with three for the harness. The benefit of the harness is therefore a lower frequency of harmful updates, with a trade-off in average performance and loss severity. Full fine-tuning performs poorly under the same small-label conditions, favoring more restricted updates for this benchmark.

The CAMELYON17 transfers make this model dependence concrete (Appendix Table~\ref{tab:cls_tasks}). At the same target hospital, the harness retains ResNet18, selects input histogram matching for Swin-Tiny, and tunes the ViT-small head. The largest gain occurs for ViT-small, whose AUROC rises from 0.703 to 0.911 through a restricted parameter update. For Swin-Tiny, a preprocessing adjustment improves an already strong pretrained model. Together, these outcomes identify distinct opportunities within one deployment setting: retaining an accurate pretrained model, matching its input distribution, and adapting its output layer. A site-wide choice of one fine-tuning recipe would overlook these differences.

Candidate choice is sensitive to which cases are annotated. Across four draws, mean AUROC is relatively stable and the selected action changes on twelve of thirteen tasks. In one NIH-to-CheXpert draw, the twelve-case selection set favors an adapted model whose test AUROC is 0.617, below the pretrained model's 0.788. This reversal shows how a small labeled set can favor an update that reduces test performance, even when average results improve.

%% file: table_classification.tex
\begin{table}[H]\centering\footnotesize
\caption{Classification with 12 labels (18 for nine-class CRC-VAL), plus a zero-label control. BN reset and Tent cover 11 transfers (pretrained-model mean 0.802 on that subset); all other rows cover 13. Harmed counts transfers with lower test AUROC after adaptation; worst $\Delta$ is the largest decrease relative to the pretrained model.}\label{tab:cls}
\begin{tabular}{lccc}\toprule
policy & mean AUROC $\uparrow$ & harmed $\downarrow$ & worst $\Delta$ $\uparrow$ \\ \midrule
source (no adaptation) & 0.810 & 0/13 & 0.000 \\
\systemname{} ($K{=}0$, label-free estimate) & 0.809 & 4/13 & $-0.082$ \\
BN reset & 0.787 & 6/11 & $-0.124$ \\
Tent & 0.779 & 7/11 & $-0.154$ \\
fixed head fine-tuning & \textbf{0.846} & 6/13 & $-0.067$ \\
fixed full fine-tuning & 0.712 & 10/13 & $-0.510$ \\
scripted ladder + Evidence Gate & 0.820 & 6/13 & $-0.183$ \\
\textbf{\systemname{}} & 0.837 & \textbf{3/13} & $-0.107$ \\
\systemname{} (maximum-performance objective) & 0.841 & \textbf{3/13} & $-0.107$ \\ \bottomrule
\end{tabular}\end{table}

%% file: appendix_supporting_tables.tex
\section{Supporting ablations and controls}

\input{table_k0_selector}

\begin{table}[H]\centering\scriptsize
\caption{Architecture dependence of BN-based adaptation. Deltas are changes from the pretrained model in AUROC (classification) or Dice (segmentation), evaluated on the full target set. Exact zeros identify actions that leave the model unchanged.}\label{tab:arch}
\setlength{\tabcolsep}{4pt}
\begin{tabular}{llrrl}\toprule
configuration & normalization & BN reset $\Delta$ & Tent $\Delta$ & interpretation \\ \midrule
ViT-small classifier & LayerNorm & 0.000 & 0.000 & no BN layers \\
Swin-Tiny classifier & LayerNorm & 0.000 & 0.000 & no BN layers \\
SegFormer 2-D & LayerNorm encoder + 1 BN head & $+0.033$ & $-0.576$ & method-dependent benefit \\
SwinUNETR 3-D~\citep{hatamizadeh2022swinunetr} & InstanceNorm + LayerNorm & 0.000 & 0.000 & no BN layers \\
SegResNet 3-D & GroupNorm & 0.000 & 0.000 & no BN layers \\
3-D U-Net, TotalSeg CE & 17 BatchNorm3d layers & $-0.373$ & $-0.114$ & both reduce Dice \\
3-D U-Net, Synapse & 17 BatchNorm3d layers & $+0.084$ & $+0.215$ & both improve Dice \\ \bottomrule
\end{tabular}
\end{table}

\begin{table}[H]\centering\footnotesize
\caption{Recoverability stress test for cross-center biparametric-MRI classification of clinically significant prostate cancer (csPCa). Fine-tuned models are evaluated on the 338 target cases outside the twelve-label adaptation set; other target rows use all 350 PCNN cases. Each reported delta uses a matched evaluation set.}\label{tab:safe_stop}
\begin{tabular}{lcc}\toprule
model and evaluation set & AUROC & $\Delta$ vs.\ target source \\ \midrule
source, in-domain RUMC ($n{=}159$) & 0.708 & n/a \\
source, target PCNN ($n{=}350$) & 0.479 & 0.000 \\
BN reset & 0.467 & $-0.012$ \\
Tent & 0.508 & $+0.029$ \\
head fine-tuning ($K{=}12$) & 0.477 & n/a \\
partial fine-tuning ($K{=}12$) & 0.501 & n/a \\ \bottomrule
\end{tabular}
\end{table}

\section{Implementation details}
Classification uses seeded, class-balanced annotation draws. Each search allows 44 controller steps and 16 candidate trials. Fixed classification fine-tuning uses AdamW, learning rate $10^{-3}$, and 80 steps, updating either the output head or all model parameters. Segmentation uses Adam and the final 40\% of parameterized leaf modules plus the head for its partial fine-tuning baseline. 

Public 3-D deployments allow 24 controller steps and a 75-minute compute budget. Search ends after 10 trials without improvement. Deployment takes 183 to 1,039 seconds, including prediction checks after reload. The Evidence Gate uses 2,000 bootstrap resamples.

%% file: table_k0_selector.tex
\begin{table}[H]
\centering
\footnotesize
\setlength{\tabcolsep}{3.6pt}
\caption{\textbf{Zero-label selector ablation on the 10 primary U-Net transfers.} The entropy baseline selects the lowest-entropy model from the same pretrained and TTA models available to the harness. Test labels are withheld from all selection rules. Decision value averages test Dice, using the pretrained-model score when no replacement is made.}
\label{tab:k0_selector}
\begin{tabular}{lccc}
\toprule
selection rule & decision value & replacements & harmful / replaced \\
\midrule
retain source & 0.483 & 0/10 & n/a \\
Minimum-entropy selector & 0.460 & 9/10 & 5/9 \\
scalar proxy $q$ & 0.460 & 8/10 & 4/8 \\
VLM preference only & 0.496 & 3/10 & 1/3 \\
\textbf{full proxy consensus} & \textbf{0.502} & 2/10 & \textbf{0/2} \\
\bottomrule
\end{tabular}
\end{table}

%% file: appendix_public_attempts.tex
\section{Public-model deployment outcomes}\label{sec:public_attempts}
All seven public-model configurations completed successfully and produced runnable model packages (Table~\ref{tab:external_public}). Five deployments selected adapted models, and two retained the pretrained model. Every package reproduced all ten test masks exactly after reload.

The seven-configuration benchmark uses the GPT-5.2 planner on one NVIDIA H200 GPU. Candidate selection requires a Dice improvement of 0.05 on the labeled selection scans. The delivered packages include the selected model, preprocessing, and inference code.

\subsection{Comparison of LLM planners}\label{sec:planner_comparison}
We compare GPT-5.2, GPT-5.6 Luna, and the open-weight Qwen3-VL-4B-Instruct and Qwen3-VL-8B-Instruct planners on the public U-Net's CT-to-AMOS MRI transfer. All four use the same checkpoint, preprocessing, annotation split, action library, adaptation seed, and four-candidate budget. Twenty annotations provide thirteen training and seven selection scans; ten separate scans measure test Dice. GPT-5.2 supplies image analysis and critique throughout. Each planner is evaluated once.

GPT-5.2 and Qwen use temperature zero and a 2,000-token response limit. Qwen uses official bfloat16 weights through Transformers 4.57.1.\footnote{Official model weights: \href{https://huggingface.co/Qwen/Qwen3-VL-4B-Instruct}{Qwen3-VL-4B-Instruct} and \href{https://huggingface.co/Qwen/Qwen3-VL-8B-Instruct}{Qwen3-VL-8B-Instruct}.} Luna uses Codex CLI 0.145.0 with low reasoning effort, default sampling, and a fresh session per decision. Its responses are accepted only within 2,000 output tokens.

\begin{table}[H]\centering\footnotesize
\caption{Planner comparison on one public-model CT-to-MRI transfer with twenty annotations. Dice is measured against expert annotations on the same ten test scans. The pretrained model has zero Dice. Reload agreement counts identical prediction masks.}\label{tab:planner_comparison}
\begin{tabular}{llcc}\toprule
planner & selected adjustment & test Dice & reload agreement \\ \midrule
GPT-5.2 & LoRA & 0.230 & 10/10 \\
GPT-5.6 Luna & LoRA & 0.196 & 10/10 \\
Qwen3-VL-4B-Instruct & LoRA & 0.215 & 10/10 \\
Qwen3-VL-8B-Instruct & partial fine-tuning & 0.205 & 10/10 \\
\bottomrule\end{tabular}
\end{table}

All four planners deliver adapted models with exact prediction agreement after reload (Table~\ref{tab:planner_comparison}). GPT-5.2, Luna, and Qwen-4B select LoRA, while Qwen-8B selects partial fine-tuning. The different scores among LoRA-based deployments highlight the role of the search and training recipe alongside the choice of adjustment family. Both open-weight planners recover segmentation from the pretrained model's zero target Dice, supporting their use within the shared adaptation workflow.